\def\usecvprtemplate{1}
\if\usecvprtemplate1
\documentclass[10pt,twocolumn,letterpaper]{article}

\usepackage[pagenumbers]{cvpr} 

\usepackage{graphicx}
\usepackage{amsmath}
\usepackage{amssymb}
\usepackage{booktabs}
\usepackage[pagebackref,breaklinks,colorlinks]{hyperref}

\usepackage[capitalize]{cleveref}
\crefname{section}{Sec.}{Secs.}
\Crefname{section}{Section}{Sections}
\Crefname{table}{Table}{Tables}
\crefname{table}{Tab.}{Tabs.}

\def\cvprPaperID{*****} 
\def\confName{CVPR}
\def\confYear{2023}

\usepackage{xspace}
\usepackage{multirow}
\usepackage{booktabs}
\DeclareMathOperator*{\argmax}{arg\,max}
\DeclareMathOperator*{\argmin}{arg\,min}

\newcommand{\etcf}{\emph{etc.\xspace}\xspace}

\newcommand{\rx}[1]{{\color{blue} [RX: #1]}}
\newcommand{\gq}[1]{{\color{green} [GQ: #1]}}
\newcommand{\zc}[1]{{\color[rgb]{1.000000,0.509804,0.278431} [ZC: #1]}}
\newcommand{\mr}[1]{{\color{red} [MR: #1]}}

\newcommand{\feat}[4]{\ensuremath{#1^{(#2,#3,#4)}}}
\newcommand{\featvec}[3]{\ensuremath{#1^{(#2,#3)}}}

\begin{document}

\title{Sanity checks and improvements for patch visualisation in prototype-based image classification}

\author{Romain Xu-Darme$^{1,2}$, Georges Quénot$^2$, Zakaria Chihani$^1$, Marie-Christine Rousset$^2$\\ \\
$^1$ Université Paris-Saclay, CEA, List, F-91120, Palaiseau, France\\
$^2$ Univ. Grenoble Alpes, CNRS, Grenoble INP, LIG, F-38000 Grenoble, France \\
{\tt\small \{romain.xu-darme, zakaria.chihani(at)cea.fr\}}\\
{\tt\small \{georges.quenot, marie-christine.rousset(at)imag.fr\}}\\
}
\maketitle
\begin{abstract}
	In this work, we perform an in-depth analysis of the visualisation methods implemented in two popular self-explaining models for visual classification based on prototypes - ProtoPNet and ProtoTree. Using two fine-grained datasets (CUB-200-2011 and Stanford Cars), we first show that such methods do not correctly identify the regions of interest inside of the images, and therefore do not reflect the model behaviour. Secondly, using a deletion metric we demonstrate quantitatively that saliency methods such as Smoothgrads or PRP provide more faithful image patches. We also propose a new relevance metric based on the segmentation of the object provided in some datasets (\eg CUB-200-2011) and show that the imprecise patch visualisations generated by ProtoPNet and ProtoTree can create a false sense of bias that can be mitigated by the use of more faithful methods. Finally, we discuss the implications of our findings for other prototype-based models sharing the same visualisation method.
\end{abstract}
\begin{abstract}
In this work, we perform an in-depth analysis of the visualisation methods implemented in two popular self-explaining models for visual classification based on prototypes - ProtoPNet and ProtoTree. Using two fine-grained datasets (CUB-200-2011 and Stanford Cars), we first show that such methods do not correctly identify the regions of interest inside of the images, and therefore do not reflect the model behaviour. Secondly, using a deletion metric we demonstrate quantitatively that saliency methods such as Smoothgrads or PRP provide more faithful image patches. We also propose a new relevance metric based on the segmentation of the object provided in some datasets (\eg CUB-200-2011) and show that the imprecise patch visualisations generated by ProtoPNet and ProtoTree can create a false sense of bias that can be mitigated by the use of more faithful methods. Finally, we discuss the implications of our findings for other prototype-based models sharing the same visualisation method.
\end{abstract}

\section{Introduction}\label{sec:intro}
\begin{figure}
\begin{subfigure}[b]{0.15\textwidth}
	\centering
	\includegraphics[width=\textwidth]{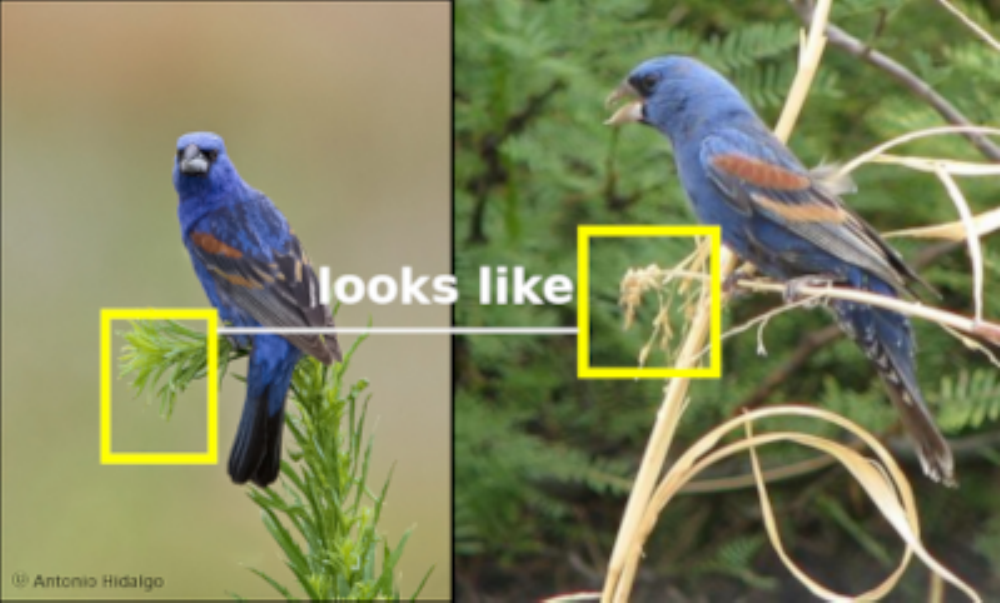}
	\caption{Upsampling.}\label{fig:inference_upsampling}
\end{subfigure}
\hfill
\begin{subfigure}[b]{0.15\textwidth}
	\centering
	\includegraphics[width=\textwidth]{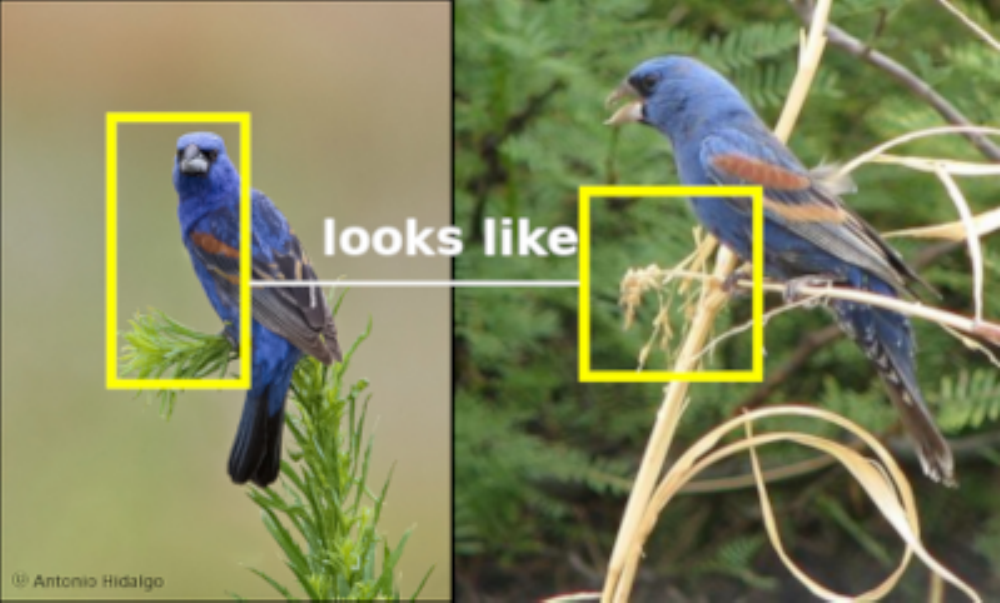}
	\caption{Smoothgrads.}
\end{subfigure}
\hfill
\begin{subfigure}[b]{0.15\textwidth}
	\centering
	\includegraphics[width=\textwidth]{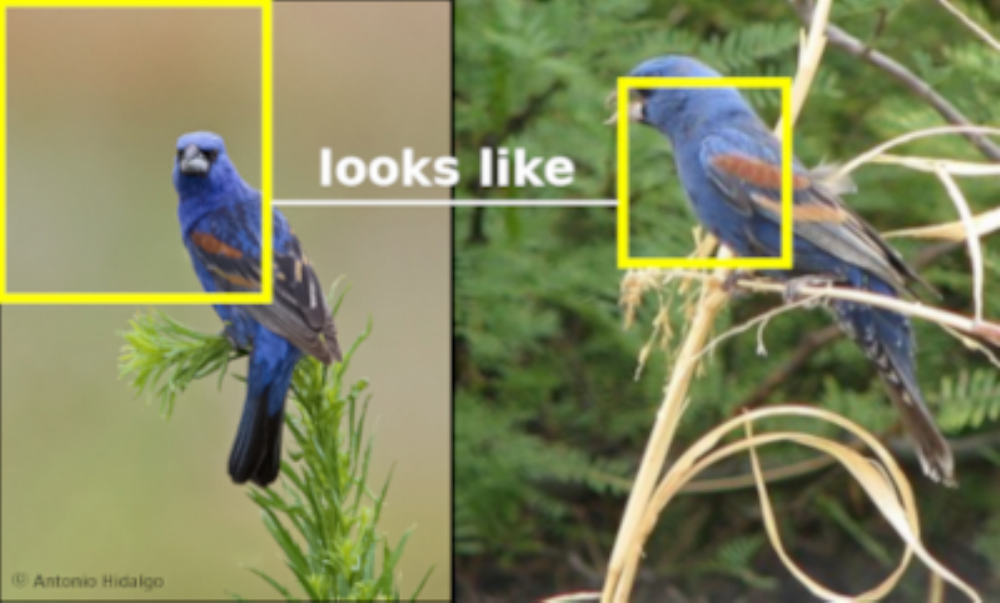}
	\caption{PRP.}
\end{subfigure}
  \caption{Explanations ("\textit{This looks like that}") during inference of a ProtoTree when using upsampling (a), Smoothgrads (b) and PRP (c) for the visualisation of image patches. Due to the imprecision of upsampling when visualising both the prototype (right) and the part in the test image (left), the user might deduce that the model is comparing tree branches, when it is actually also taking the bird into account.}
  \label{fig:comparisons}
\end{figure}

During the last decade, the field of Explainable AI (XAI) has progressively gained wide-spread recognition among the scientific community~\cite{adadi2018peeking,arrieta2019explainable,bodria2021benchmarking, nauta2022anecdotal}. This evolution reflects the growing need by society for more transparent and accountable systems, especially in critical domains (\eg autonomous driving, medical diagnosis), as demonstrated by the recent adoption of the European General Data Protection Regulation (GDPR) and the proposal for a European AI Regulation Act.\\
XAI aims at unravelling the decision-making process of an autonomous system, and encompasses two approaches that are sometimes conflicting~\cite{rudin2019stop}, yet often complementary. \textit{Post-hoc} explanation methods apply to pre-existing models (\ie pre-trained models in the context of machine-learning) and provide various information regarding either a particular decision (\textit{local explanation}) or the general model behaviour (\textit{global explanation}): understandable local approximation~\cite{ribeiro2016why,lundberg2017unified}, saliency maps~\cite{simonyan2013deep,shrikumar2016not,sundararajan2017axiomatic,smilkov2017smoothgrads,binder2016layer},
counterfactuals~\cite{wachter2017counterfactual, laugel2018comparison,looveren2019interpretable, dhurandhar2019model}, concept-based explanations~\cite{kim2017interpretability,wu2020towards}, \etcf In particular, \textit{post-hoc} explanation methods can help explain the decision-making process of Deep Neural Networks (DNNs), which are highly efficient yet highly complex systems. In the field of computer vision for instance, such methods can help identify the attention region of the DNN \wrt a particular decision, sometimes highlighting a possible bias in the model~\cite{ribeiro2016why}. However, it is worth mentioning that this attention region does not - by itself - necessarily  explain the model's decision. Indeed, a DNN might be focusing on a relevant part of the image, and yet might produce an incorrect decision, based on high-dimensional features that are inherently abstract due to the network training procedure.
\\
In order to mitigate this limitation in the explainability of complex systems such as DNNs, a second major avenue of research in the field of XAI consists in  developing architectures and training procedures such that the resulting model should be \textit{explainable-by-design} (such models are also called \textit{self-explaining} or \textit{interpretable-by-design}). In computer vision, such architectures primarily use either a cased-based reasoning mechanism~\cite{chen2019this,nauta2021neural,rymarczyk2021protopshare,rymarczyk2021interpretable,donnelly2021deformable} - where new instances of a problem are solved using comparison with examples (\textit{prototypes}) extracted from the training dataset - or concept-based attribution~\cite{alvarezmelis2018towards,chen2020concept}. In particular, ProtoPNet~\cite{chen2019this} and ProtoTree~\cite{nauta2021neural} have shown that explainable-by-design architectures can reach performance levels on par with non-interpretable models on fine-grained recognition tasks~\cite{welinder2010caltech,yang2015largescale}. During training, both models extract reference vectors in the latent space of a deep convolutional neural network (DCNN), corresponding to patches of images in the training set (the so-called \textit{prototypes}). During inference, the similarity between a test image and a prototype is evaluated by computing the distance between their respective representations in the latent space of the DCNN. Finally, ProtoPNet and ProtoTree base their decision using the prototypes that show the highest similarity with the test image, and produce explanations by displaying patches of the test image and their most "similar" prototypes side-by-side (\textit{this looks like that}~\cite{chen2019this}). \\
However, in practice, both ProtoPNet and ProtoTree sometimes provide explanations using image patches that seem to be focused on the background or elements unrelated to the object itself, indicating a possible bias in the decision (see Fig.\ref{fig:inference_upsampling}).
Moreover, while a patch from the test image focusing on the background might invalidate one particular inference, a prototype focusing on background might indicate a more \textit{systemic} bias in the model and seriously hinder its practical accuracy. Although such issue may undermine the trust in self-explaining models, there exists in reality a fundamental imprecision in the patch visualisation methods implemented by ProtoPNet and ProtoTree to generate explanations. In particular, \cite{gautam2022this} has shown that this imprecision in ProtoPNet visualisation method can sometimes hide a specific type of bias - known as the "Clever Hans" phenomenon - where a given class is highly correlated to a specific visual artefact unrelated to the task at hand (\eg watermarking in an image). More generally, imprecise visualisation methods may suggest model bias where there is none, while sometimes hiding systemic issues of the model.
\paragraph{Our contribution:} In this work, we perform an analysis of the visualisation methods implemented in ProtoPNet and ProtoTree, attempting to answer the following research questions: do these architectures generate faithful image patches corresponding to the latent representations used in their decision-making process? do they produce decision based on relevant parts of the image or based on elements of the background? Using two fine-grained datasets (CUB-200-2011~\cite{welinder2010caltech} and Stanford Cars~\cite{yang2015largescale}) and two saliency methods (Smoothgrads~\cite{smilkov2017smoothgrads} and PRP~\cite{gautam2022this}), we confirm the results of \cite{gautam2022this} on ProtoPNet and show that ProtoTree also generates imprecise visual patches. Additionally, using the object segmentation provided in the CUB-200-2011 dataset, we propose a new relevance metric and show quantitatively that in both architectures, such imprecise visualisations often create a false sense of bias that is largely mitigated by the use of more faithful methods. Finally, we discuss the implications of our findings to other prototype-based models sharing the same visualisation method.\\
This paper is organized as follows: Section~\ref{sec:related_work} describes the related work; Section~\ref{sec:theory} recalls the theoretical background of ProtoPNet and ProtoTree and introduces the metrics used to evaluate the fidelity and relevance of prototype-based explanation; Section~\ref{sec:experiments} describes the results of our experiments. Finally, Section~\ref{sec:discussion} concludes this contribution and proposes possible improvements to self-explaining models.

\section{Related work}\label{sec:related_work}
\paragraph{Prototype-based classifiers}
Due to the difficulty of quantifying similarity between images in the visual space (\ie the space of RGB images), self-explaining image classification models based on prototypes first encode images into a high dimensional feature space (also called latent space) - usually using a pre-trained DCNN (\eg Resnet50~\cite{he2016residual}, VGG~\cite{simonyan2015very}) called a backbone, assuming that such encoding preserves visual cues (\eg colour, shape) while being insensitive to minor deformations such as rotations, shifts, changes in scale. This latent representation can be composed of a single vector summarizing the information of the entire image, or composed of an array of vectors - corresponding to the output of a convolution layer - and retaining a form of spatial relationship between each vector and a region of the input image (see Fig.~\ref{fig:this_looks_like_that}). Since this spatial relationship is at the heart of this contribution, we purposefully exclude self-explaining image classification models that uses full images as prototypes~\cite{arik2019protoattend, angelov2019towards, angelov2020towards} from our study. Indeed, such architectures use a single latent vector representation, resulting in a trivial relationship between the full image and its latent representation. In this contribution, we rather focus on models where each vector in the latent space corresponds to \textit{a part} of an image. During training, such models extract a set of reference vectors and their visual counterparts from the training set, called \textit{part prototypes} (for simplicity, we simply use the term \textit{prototypes}). Prototypes are either discriminative of a particular class~\cite{chen2019this,donnelly2021deformable,wang2021interpretable} (given by the label of the corresponding training image) or shared among multiple classes~\cite{nauta2021neural,rymarczyk2021interpretable,rymarczyk2021protopshare}.
During inference, the similarity between a given prototype and the test image is computed using the L2-distance (or cosine distance~\cite{wang2021interpretable}) between their respective latent representations, under the assumption that proximity between vectors in the latent space should entail similarity in the visual space. In the case of ProtoPNet~\cite{chen2019this}, ProtoPShare~\cite{rymarczyk2021protopshare}, ProtoPool~\cite{rymarczyk2021interpretable}, Deformable ProtoPNet~\cite{donnelly2021deformable} and TesNet~\cite{wang2021interpretable},
all similarity scores are then processed through a fully connected layer to produce the final classification. In the case of ProtoTree~\cite{nauta2021neural}, these similarity scores are used to compute a path across a soft decision tree where each leaf corresponds to one particular class. \\
It is important to indicate that although our study focuses on ProtoPNet and ProtoTree, all the aforementioned methods (ProtoPShare, ProtoPool, Deformable ProtoPNet, TesNet) share a common code base inherited from ProtoPNet and therefore are theoretically susceptible to the issue raised in this contribution.

\paragraph{Saliency methods} Such methods are among the most commonly used building blocks for generating explanations and aim at identifying the most important (salient) features (pixels) of an image \wrt the output of a given neuron. Although they are usually used to explain the decision taken by the model (\eg neurons from the last layer of a classifier), they can also be used to visualise the most salient part of the image \wrt any intermediate result of a DCNN. Gradient-based approaches~\cite{simonyan2013deep} compute the partial derivative $\frac{\delta S}{\delta x_i}$ of the target neuron output $S$ \wrt to each input pixel $x_i$, with improvements such as Integrated Gradients~\cite{sundararajan2017axiomatic} or SmoothGrads~\cite{smilkov2017smoothgrads} aiming at generating more stable explanations. In particular, Smoothgrads "adds noise to remove noise" by averaging gradients over noisy copies $x + n_i$ ($n_i \sim \mathcal{N}(0,\sigma^2)$) of $x$. Note that in order to take into account the sign and strength of the input~\cite{shrikumar2017learning}, it is also possible to perform an element-wise product between the gradient and the input ($\frac{\delta S}{\delta x_i} \odot x$). Layer-wise Relevance Propagation~\cite{binder2016layer} (LRP) uses a set of rules in order to back-propagate the relevance of each individual neuron from the target back to the input, following a conservation property. This method has several variants depending on the applied rules~\cite{montavon2019layer}. In the particular context of prototype-based models, \cite{gautam2022this} proposes a variant of LRP called Prototype Relevance Propagation (PRP), implementing a dedicated rule to propagate relevance across the layer in charge of computing similarity scores, followed by LRP$_{COMP}$~\cite{kohlbrenner2019towards}. Interestingly, \cite{ancona2017unified} and \cite{shrikumar2016not} have shown that, for DCNNs based on ReLU activation (as it is often the case for the networks used for feature extraction in self-explaining models), Integrated Gradients, LRP with $\epsilon$-rule and Deep-LIFT~\cite{shrikumar2017learning} are all equivalent to $\frac{\delta S}{\delta x_i} \odot X$. For these reasons, in this work, we choose to compare the original part visualisation generated by ProtoPNet and ProtoTree to visualisations generated using Smoothgrads $\odot$ input and PRP.
Note that we also exclude Guided-Backpropagation~\cite{springenberg2014striving} and its application to GRADCAM~\cite{selvaraju2016grad} due to the results of \cite{adebayo2018sanity} which raise some concerns regarding the relevance of the explanations based on these methods. \\
We emphasize the fact that a saliency map \textit{is not an explanation by itself}, but is rather used to build an explanation. In the context of self-explaining models based on prototypes, the saliency maps of both the prototype and the test image are cropped in order to retain only the most salient pixels in the corresponding images. The explanation finally consists in the parallel display of the selected regions in the two images.

\paragraph{Evaluation metrics} Explanations based on part visualisation implicitly rely on the underlying method used for generating saliency maps. Multiple metrics have been proposed in order to evaluate the quality of explanations against a set of desired properties~\cite{nauta2022anecdotal}. In order to answer our first research question (do ProtoPNet and ProtoTree generate faithful image patches corresponding to the latent representations used in their decision-making process?), we focus on the property known as \textit{faithfulness}~\cite{alvarezmelis2018towards} (\textit{a.k.a.} \textit{fidelity}~\cite{tomsett2019sanity} or \textit{correctness}~\cite{nauta2022anecdotal}) which - in our case - quantifies the adequacy between a saliency method and the model behaviour. In this particular context, faithfulness can primarily be evaluated by model parameter randomisation or deletion/insertion methods. Parameter randomisation~\cite{adebayo2018sanity} analyses the effect of perturbations to the model parameters on the saliency map: if a method produces the same saliency maps regardless of the changes in the model parameters, then it probably does not correctly reflect the model behaviour. As described above, in this regard the results obtained by~\cite{adebayo2018sanity} guide us in our choice of saliency methods. Deletion/insertion metrics~\cite{petsiuk2018rise,alvarezmelis2018towards,tomsett2019sanity} monitor the evolution of the target neuron's output when the most/least important pixels of the input image are removed incrementally~\cite{petsiuk2018rise,tomsett2019sanity} or individually~\cite{alvarezmelis2018towards}. Such metrics aim at checking the ability of a saliency method to correctly sort the importance of pixels \wrt to a particular model output. In particular, "removing" the most salient pixels identified by a saliency method faithful to the model should result in a strong variation of the target neuron's output, while a less faithful method will highlight pixels whose absence after deletion has actually no influence on the neuron's output. In this work, we choose to implement an incremental deletion metric, limiting ourselves to 2\% of the original image in order to avoid evaluating our saliency methods using out-of-distribution inputs~\cite{gomez2022metrics}. \\
Finally, \cite{nauta2020this} evaluates the \textit{relevance} of prototype-based explanations by applying controlled perturbations (changes in colour hue, shape, texture, saturation, contrast) on images and monitoring the evolution of the similarity score for each prototype. Note that since these perturbations are applied on the \textit{entire} image, this method is incidentally independent of the spatial location of the target image patch.
In this work, we wish to check two properties of ProtoPNet and ProtoTree that are related to our second research question: after training, does each prototype correspond to a part of the object? during inference, do the patches from the test image that are compared to the prototypes also correspond to parts of the object?
Therefore, we propose to focus on the intersection between the pixels highlighted by the saliency methods and the ground-truth segmentation of the object (when available). We do not use the popular Intersection-over-Union (IoU) metric because our goal is not to assert whether each prototype covers the entire object, but rather measure the percentage of the image patch intersecting the object segmentation.

\section{Theoretical background}\label{sec:theory}
In this work, we consider a classification problem with a training set $X_{train} \subset \mathcal{X}\times \mathcal{Y}$. Let $f$ be a fully convolutional neural network (fCNN) processing images in $\mathcal{X}$ and producing a $D$-dimensional latent representation of size $(H,W)$. We denote $\mathcal{L}_f=\mathbb{R}^{H\times W\times D}$ the latent space
associated with $f$. For $x\in \mathcal{X}$, we denote $\featvec{f}{h}{w}(x)\in \mathbb{R}^D$ the $D$-dimensional vector corresponding to the $h$-th row and $w$-th column of $f(x)$.

\subsection{ProtoPNet and ProtoTree}
\paragraph{Finding similarities in the latent space}
For $x\in \mathcal{X}$, ProtoPNet and ProtoTree compute their decision (classification) based on similarities between the latent representation $f(x)$ and a set of $D$-dimensional vectors $(r_1, \ldots r_p)$ that are learned during training and act as reference points in the latent space. More precisely, the similarity between $r_i$ and a particular vector $\featvec{f}{h}{w}(x)$ is defined as
$\featvec{s_i}{h}{w}(x)= log\big((\|\featvec{f}{h}{w}(x)-r_i\|_2^2+1)/(\|\featvec{f}{h}{w}(x)-r_i\|_2^2+\epsilon)\big)$ (ProtoPNet) or $\featvec{s_i}{h}{w}(x)=e^{-\|\featvec{f}{h}{w}(x)-r_i\|_2^2}$ (ProtoTree), where $\|.\|_2$ denotes the L2 distance.
For each reference point $r_i$, $s_i(x)\in \mathbb{R}^{H \times W}$ is called the \textit{similarity map} between $x$ and $r_i$.
The model decision process $d$ is a function of the aggregation of high similarity scores $s(x)=(\max(s_1(x)), \ldots \max(s_p(x)))$: weighted sum for ProtoPNet, soft decision tree for ProtoTree. During training, the parameters (convolutional weights) of the feature extractor $f$, of the decision function\footnote{The details of the decision function $d$ are not relevant to our work, which focuses on the method used by ProtoPNet and ProtoTree to build a saliency map out of the similarity map $s_i(x)$.} $d$, and the reference points $r_i$ are jointly learned in order to minimize the cross-entropy loss between the prediction $d \circ s(x)\in \mathcal{Y}$ and $y$, $\forall (x,y)\in X_{train}$.

\paragraph{Prototype projection}
After training, the reference points $r_i$ are "pushed" toward latent representations of parts of training images, in a process called \textit{prototype projection}. More formally, a prototype $P_i$ is a tuple $P_i=(p_i, h_i, w_i, r_i)$, computed using a projection dataset $X_{proj} \subset \mathcal{X}$ (usually consisting of images from the training set $X_{train}$), where
\begin{equation}
	\left\{
	\begin{array}{l}
		p_i, h_i, w_i = \argmin\limits_{x\in X_{proj}, h, w} \|\featvec{f}{h}{w}(x)-r_i\|_2^2 \\
		r_i \leftarrow \featvec{f}{h_i}{w_i}(p_i)
	\end{array}
	\right.
\end{equation}
Thus, prototype projection moves each reference point - corresponding to an abstract vector in the latent space - to a nearby point which is, by construction, obtained from an image $p_i$. Although this operation may reduce the accuracy of the system (changing the value of the reference points $r_i$ may increase the cross-entropy loss), it allows the system to anchor the reference points as latent representations of real images from the projection set.

\begin{figure}
    \centering
    \includegraphics[width=\linewidth]{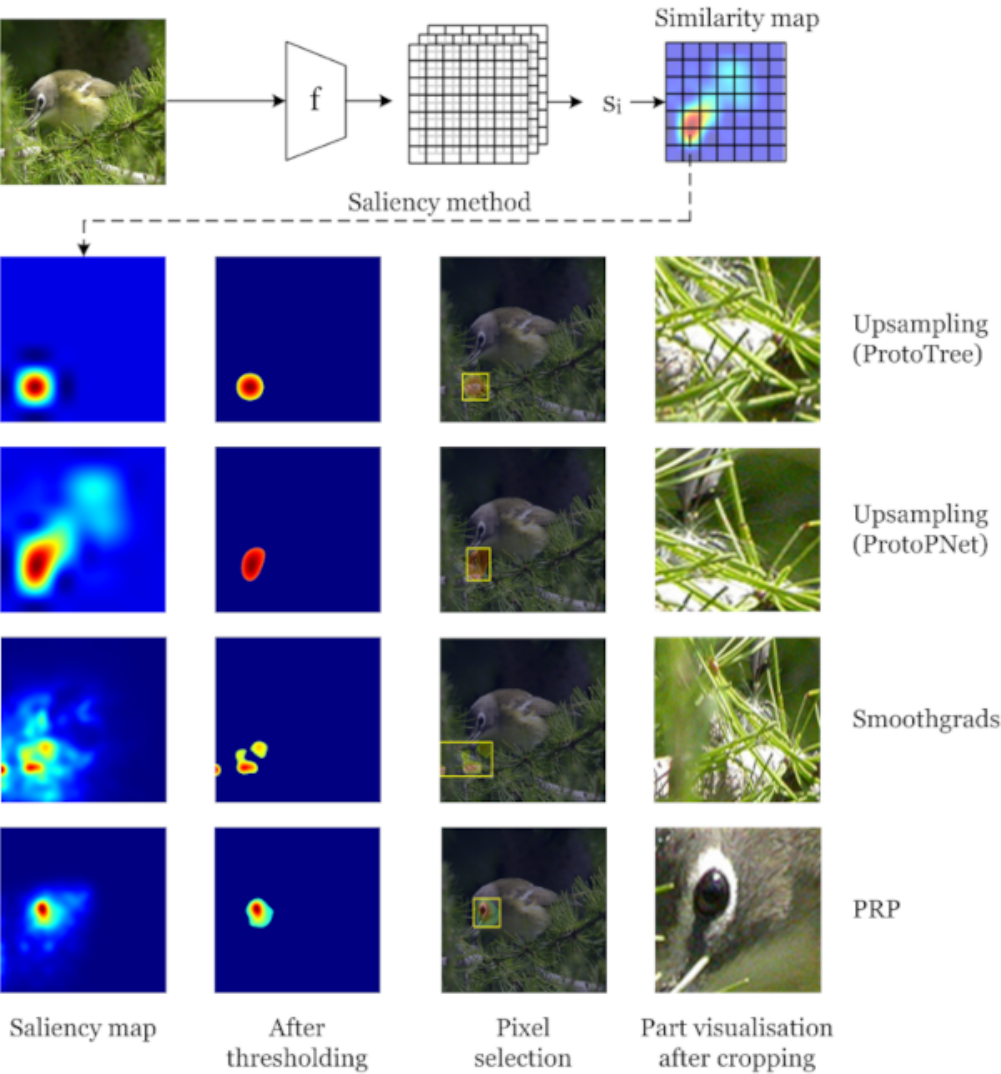}
    \caption{Generating part visualisation from the similarity map. First, a saliency method generates a saliency map from the similarity map. Then, thresholding is applied to retain only the most salient pixels. Finally the original image is cropped to produce a part visualisation.}
    \label{fig:part_visualisation}
\end{figure}

\paragraph{From similarity map to part visualisation}
Given an image $x$ and a prototype $P_i$, ProtoPNet generates a visualisation of the most similar image patch in $x$ by upsampling the similarity map $s_i(x)\in \mathbb{R}^{H\times W}$ to the size of $x$ using cubic interpolation, then cropping the resulting saliency map to the 95th percentile, while ProtoTree retains only the location of highest similarity in $s_i(x)$ before upsampling (setting all other locations to $0$), as shown in Fig.~\ref{fig:part_visualisation}. Note that the same method is also used to visualise the prototype itself by setting $x=p_i$ and that prototype visualisation can be performed once and for all after projection.
However, both approaches do not factor in the size of the receptive field\footnote{This information is actually computed in the code of ProtoPNet, but never put to use.} of each neuron in the last layers of a deep CNN. Indeed, taking into account padding and pooling layers in the architecture of DCNNs, the value $\featvec{s_i}{h}{w}(x)$ may actually depend on the entire image $x$~\cite{araujo2019computing} rather than a localized region.

\begin{figure}
	\centering
	\includegraphics[width=\linewidth]{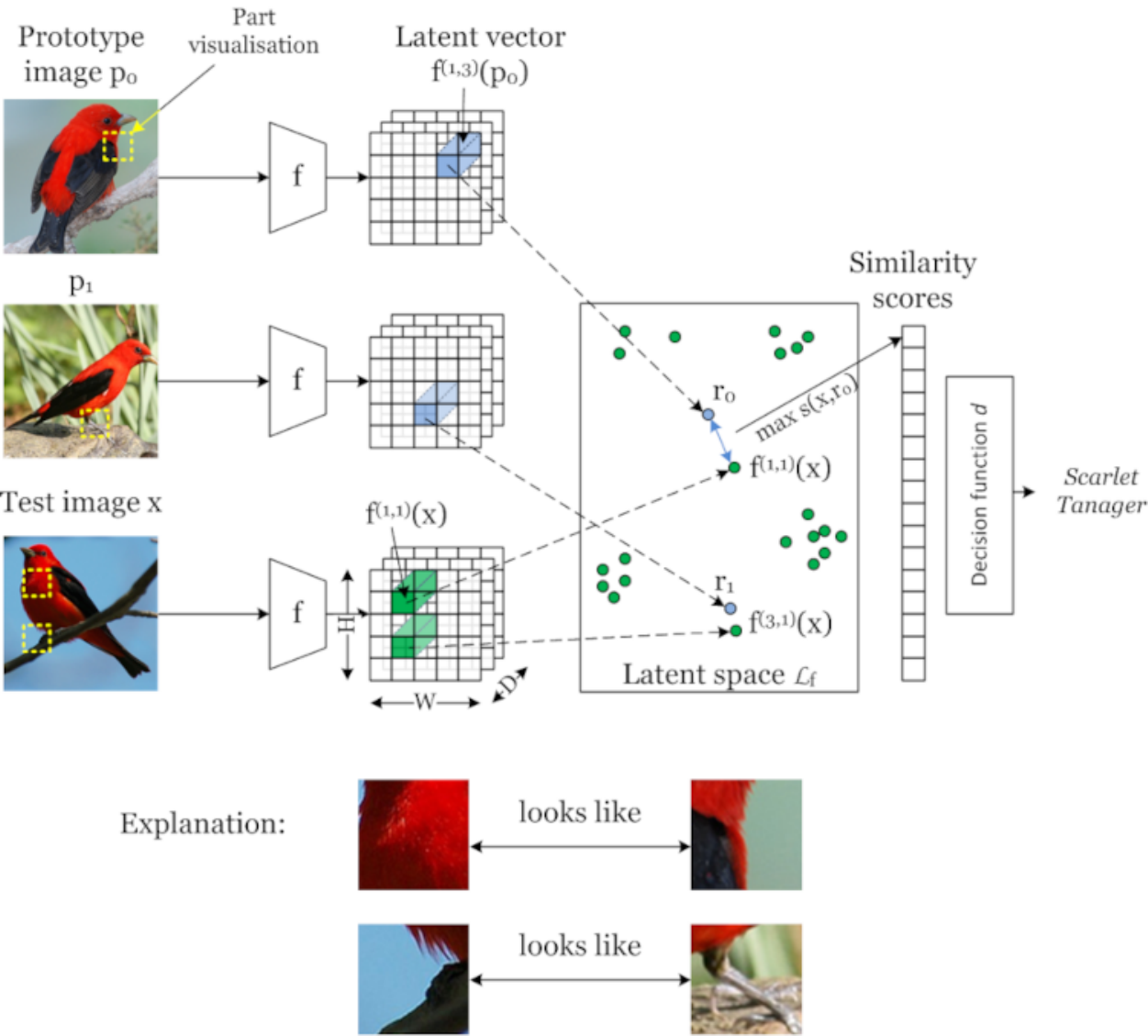}
	\caption{ProtoPNet/ProtoTree inference overview. After training and projection, a prototype $P_i$ is composed of an image from the training set ($p_i$) and a latent vector (\eg $r_0=\featvec{f}{1}{3}(p_0)$). During inference, similarity scores are computed based on the closest distance to each prototype latent representation and used in the decision function $d$ to produce the prediction. Then, part visualisation is applied to associate each latent vector to a patch inside of its original image. Finally, an explanation is generated by visualising side by side all relevant prototypes and their most similar image patches in $x$. Best viewed in colour.}
	\label{fig:this_looks_like_that}
\end{figure}

\paragraph{\textit{This} looks like \textit{that}}
As illustrated in Fig.~\ref{fig:this_looks_like_that}, during inference, for an image $x\in \mathcal{X}$, both ProtoPNet and ProtoTree base their decision on the comparison between the latent representation $f(x)$ and the representatives of the prototypes. More precisely, for each prototype $P_i$, the model finds the vector in $f(x)$ closest to $r_i$, corresponding to the highest score of the similarity map $s_i(x)$. If this score is above a given threshold, then it generates and shows side by side patches of images extracted from the prototype image $p_i$ and the test image $x$.

\subsection{Generating part visualisation with PRP and Smoothgrads}
Similar to ProtoTree, for an image $x\in \mathcal{X}$ and a prototype $P_i$, we first find the coordinates and value of the highest similarity score
\begin{equation}
    \left\{
	\begin{array}{l}
        h_m,w_m = \argmax\limits_{h,w} \featvec{s_i}{h}{w}(x) \\
        s_m = \max\limits_{h,w} \featvec{s_i}{h}{w}(x)=\featvec{s_i}{h_m}{w_m}(x)
	\end{array}
	\right.
\end{equation}
Where ProtoPNet and ProtoTree directly upsample the similarity map to the size of the original image $x$, in this work we generate saliency maps identifying the most important pixels \wrt the highest similarity score $s_m$ by applying  Smoothgrads or PRP on the output of the neuron $\featvec{s_i}{h_m}{w_m}$. Importantly, we perform a back-propagation of the similarity score \textit{through} the feature extractor $f$ in order to take into account its architecture and parameters. Then, we obtain a part visualisation by using the saliency map to retain only the 2\% of most important pixels from $x$ and cropping the image accordingly, as shown in Fig.~\ref{fig:part_visualisation}. Again, the same method is applied in order to extract a part visualisation for each prototype.

\subsection{Measuring faithfulness}\label{sec:faithfulness}
\begin{figure}
	\centering
    \includegraphics[width=\linewidth]{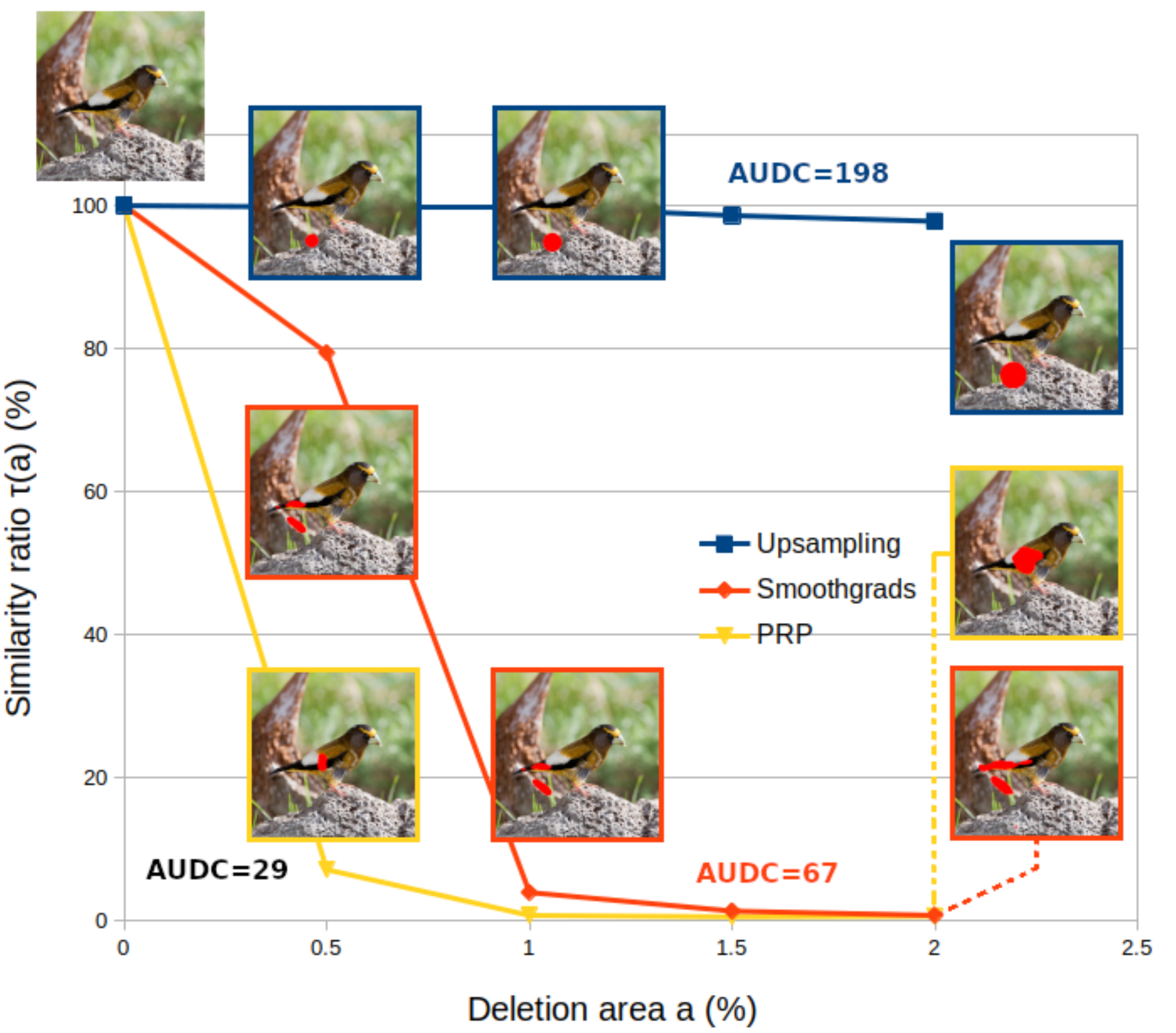}
     \caption{Evolution of the similarity ratio when incrementally removing the most important pixels according to the ProtoPNet/ProtoTree method (upsampling, in blue), Smoothgrads (red), and PRP (yellow). In this example, removing pixels according to upsampling has little to no effect on the similarity score, suggesting that the explanation is incorrect. On the contrary, when removing only 1\% of the image according to Smoothgrads, the similarity score drops to 2\% of its original value, suggesting that the explanation focuses on actual regions of interest for the model. The same result is achieved when removing only 0.5\% of the image according PRP, indicating an even more precise explanation. Moreover, reaching a similarity ratio lower than 10\% indicates that the explanation method has successfully identified the most relevant pixels of the image patch and gives an indication on the effective size of the image patch. Best viewed in colour.}
     \label{fig:deletion_curve}
\end{figure}
\begin{figure}
	\centering
    \includegraphics[width=0.8\linewidth]{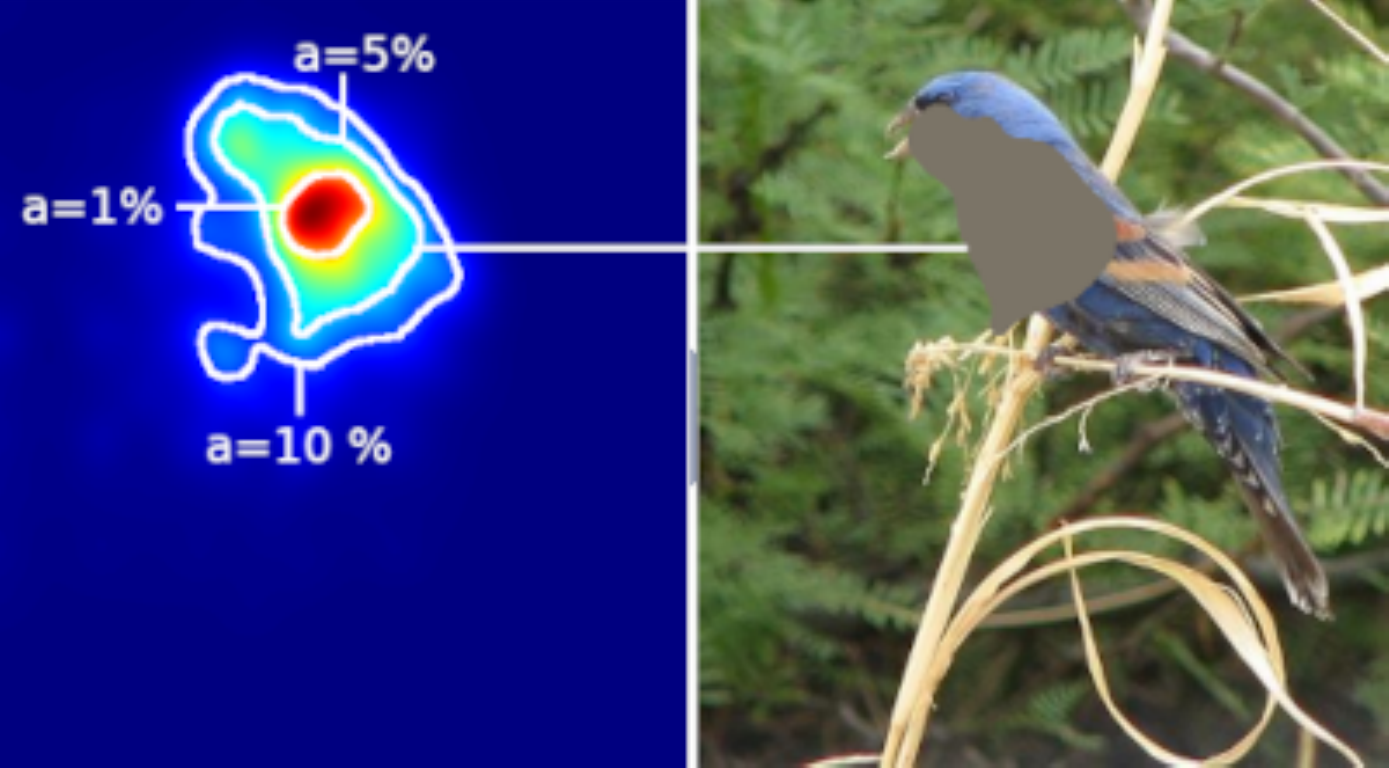}
    \caption{From a saliency map generated by a saliency method (left), we progressively mask out a growing area of image $x$, selecting the most salient pixels first, and generate perturbed images $\tilde{x}$ (right). By studying the impact of perturbed images on the similarity score, the deletion metric evaluates the faithfulness of the saliency method.}\label{fig:saliency_mask}
\end{figure}

As illustrated in Fig.~\ref{fig:deletion_curve}, and similar to \cite{petsiuk2018rise}, we measure the faithfulness of a saliency method to the model behaviour by analysing the drop in similarity when incrementally "deleting" pixels of highest relevance.
In practice, for an image $x\in \mathcal{X}$, a prototype $P_i$ and a given saliency method (upsampling, Smoothgrads, PRP), we first generate the saliency map from the similarity map $s_i(x)$. Then, for a target deletion area $a$, we measure the drop in similarity as follows:
\begin{itemize}
    \item We generate an image $\tilde{x}$ obtained after masking out the $a$\% most salient pixels from $x$ (see Fig.~\ref{fig:saliency_mask}).
    \item We compute the value $\tilde{s}_m = \featvec{s}{h_m}{w_m}(\tilde{x})$ corresponding to the similarity score of $\tilde{x}$ at the original location of highest similarity.
    \item We compute the value $\tau(a)=\tilde{s}_m/s_m$ measuring the relative drop in similarity between $x$ and $\tilde{x}$.
\end{itemize}

$\tau(a) \approx 1$ indicates that deleting these pixels has no impact on the similarity score, thus that the saliency method is not faithful to the model. On the contrary, $\tau(a) \approx 0$ indicates that deleting these pixels has a high impact on the similarity score, thus that the saliency method correctly identifies relevant pixels \wrt to the similarity score and is faithful to the model. Finally, we compare the faithfulness of all saliency methods under test by computing the Area under the Deletion Curve (AUDC) for values of $a$ up to 2\% of the total image area (as indicated in Sec.~\ref{sec:related_work}, we restrict ourselves to deleting a small portion of the original image in order to avoid unexpected behaviours from the DCNNs~\cite{gomez2022metrics}).
Interestingly, identifying the minimum deletion area leading to a significant drop in similarity also provides an indication regarding the size of the \textbf{effective receptive field} of the model \wrt to a given neuron. Indeed, as shown in Fig.~\ref{fig:deletion_curve}, reaching a similarity ratio $\tau(a) < 0.1$ indicates that the deleted pixels amount to 90\% of the similarity score, thus that the area of the effective receptive field is close to $a$.

\subsection{Measuring relevance}\label{sec:relevance}

In order to verify that the model is producing decisions based on parts of the object rather than elements of the background,
we measure the relevance of both prototypes and their most similar patches in test images. As mentioned in Sec.~\ref{sec:related_work}, we measure the relevance of an image patch as its intersection with the object segmentation, assuming that such information is available in the training data. In practice, for a given saliency method, an input image $x$ and a prototype $P_i$, we first identifies the 2\% most salient pixels of $x$ \wrt the maximum similarity score. If few or none of these pixels (less than 5\% in practice) intersect the object segmentation, then the image patch is considered irrelevant, as it mostly focuses on the background. As shown in Fig.~\ref{fig:comparisons}, different methods lead to different saliency maps that produce image patches with different relevance. The effective relevance of prototypes and patches from the test image is therefore dependent on the faithfulness of the chosen saliency method.

\section{Experiments}\label{sec:experiments}
\paragraph{Setup}
We perform our experiments on two popular fine-grained datasets: the CUB-200-2011~\cite{welinder2010caltech} dataset (CUB) contains 11,788 images belonging to 200 bird species - split into 5,994 training images and 5,794 test images - and provides the object bounding box coordinates and segmentation mask for each image; the Stanford Cars~\cite{yang2015largescale} dataset (CARS) contains 16,185 images belonging to 196 car models, evenly split into training and test images.
Additionally, for ProtoPNet, we use the cropped images of the CUB dataset - which we denote CUB-c - during training and inference.
As summarized in Table~\ref{tab:accuracies}, for both models, we primarily use a Resnet50~\cite{he2016residual} feature extractor (\textit{backbone}), pretrained on the iNaturalist~\cite{horn2017inaturalist} dataset (CUB) or the ImageNet~\cite{deng2009imagenet} dataset (CARS), with images of size $224\times 224$. In order to compare results across different feature extractors, we also train a ProtoPNet on CUB using a VGG19~\cite{simonyan2015very} network (pretrained on ImageNet).

\begin{table}[]
\centering
\small
	\begin{tabular}{|c|c|c|c|}
		\hline
		Model & Backbone & Dataset & Accuracy \\
		\hline
		\multirow{3}{*}{ProtoPNet} & VGG19    & CUB-c & 75.1\% \\
		\cline{2-4}
                                   & \multirow{2}{*}{ResNet50} & CUB-c & 72.5\% \\
                                                  	         & & CARS & 71.4\% \\
        \hline
        \multirow{2}{*}{ProtoTree} & \multirow{2}{*}{ResNet50} & CUB & 83.1\% \\
                                                  	         & & CARS & 83.2\%\\
        \hline
	\end{tabular}
	\caption{Accuracy of the self-explaining models used in this work. CUB-c denotes the cropped CUB-200-2011 dataset.}\label{tab:accuracies}
\end{table}

For saliency methods, we use the code of PRP~\cite{gautam2022this} kindly provided by the authors. For Smoothgrads~\cite{smilkov2017smoothgrads}, we use 10 noisy samples per image and dynamically set the value of $\sigma$  using a noise ratio of $0.2$ (\ie $\sigma = 0.2 \times\Delta x$). For both methods, we post-process the saliency map by successively averaging the gradients at each pixel location across the RGB channels, taking the absolute value (putting equal emphasis on strongly positive and negative gradients), and applying a $5\times 5$ Gaussian filter in order to avoid isolated gradients due to max-pooling layers inside of $f$.

\paragraph{Faithfulness of patch visualisation}
\begin{table}[]
\centering
\scriptsize
	\begin{tabular}{|c|c|c|ccc|}
		\hline
		\multirow{2}{*}{Model} & \multirow{2}{*}{Backbone} & \multirow{2}{*}{Dataset} & \multicolumn{3}{c|}{Method} \\
		& & & Up. & PRP & S $\odot$ I \\
		\hline
		\multirow{3}{*}{ProtoPNet} & VGG19    & CUB-c & 82/148 & 77/139 & \textbf{74/136}\\
		\cline{2-6}
                                   & \multirow{2}{*}{ResNet50} & CUB-c & 78/142 & \textbf{62/121} & 74/132 \\
                                                  	         & & CARS & 91/176 & \textbf{61/136} & 68/141 \\
        \hline
        \multirow{2}{*}{ProtoTree} & \multirow{2}{*}{ResNet50} & CUB & 196/190 & \textbf{153/130} & 181/164 \\
                                                  	         & & CARS & 192/181 & \textbf{150/142} & 175/168 \\
        \hline
	\end{tabular}
	\caption{Average Area Under the Deletion Curve (AUDC$\downarrow$) of prototypes (left value) and test patches (right value) generated by ProtoPNet and Prototree when using upsampling, Smoothgrads ($S\odot I$) and PRP. For readability, all values are multiplied by a factor 10,000, with AUDC=200 corresponding to a similarity ratio $\tau$ maintained at 100\% for all deletion areas up to 2\% (see Fig.~\ref{fig:deletion_curve}).}\label{tab:faithfulness}
\end{table}
We evaluate the faithfulness of the saliency methods under test by measuring the AUDC (as described in Sec.~\ref{sec:faithfulness}) when visualising prototypes after training, but also on patches of images from the test set during inference:
for ProtoTree, we apply the saliency method only when a positive comparison is established between a patch of the test image and a prototype (right branch of each decision node); for ProtoPNet, we apply the saliency method on the 10 patches of the test image that are most similar to any prototype of the inferred class. The AUDC score is measured by using deletion areas between 0\% and 2\%, with an increment value of 0.1\%.
As shown in Table~\ref{tab:faithfulness}, in all cases, the upsampling method used in ProtoPNet and ProtoTree leads to a higher AUDC score than Smoothgrads or PRP. This confirms the imprecision pointed out in \cite{gautam2022this} and \textit{extends the issue to ProtoTree}. Moreover, except in the case of ProtoPNet on CUB-c using a VGG19 backbone, PRP seems to provide a more faithful saliency maps than Smoothgrads (lower AUDC), at the cost of a higher computing time due to its rules for relevance propagation (on a Nvidia Quadro T2000 Mobile, a single PRP relevance propagation is approximately 1.4 times slower than Smoothgrads with 10 noisy samples). However, we note that the AUDC scores between upsampling, Smoothgrads and PRP are fairly similar on the CUB-c dataset, which is probably due to the image cropping that increases the relative area of the bird inside of the image and decreases the probability to miss the important pixels, even with a random guess.

\begin{figure}
    \centering    \includegraphics[width=\linewidth]{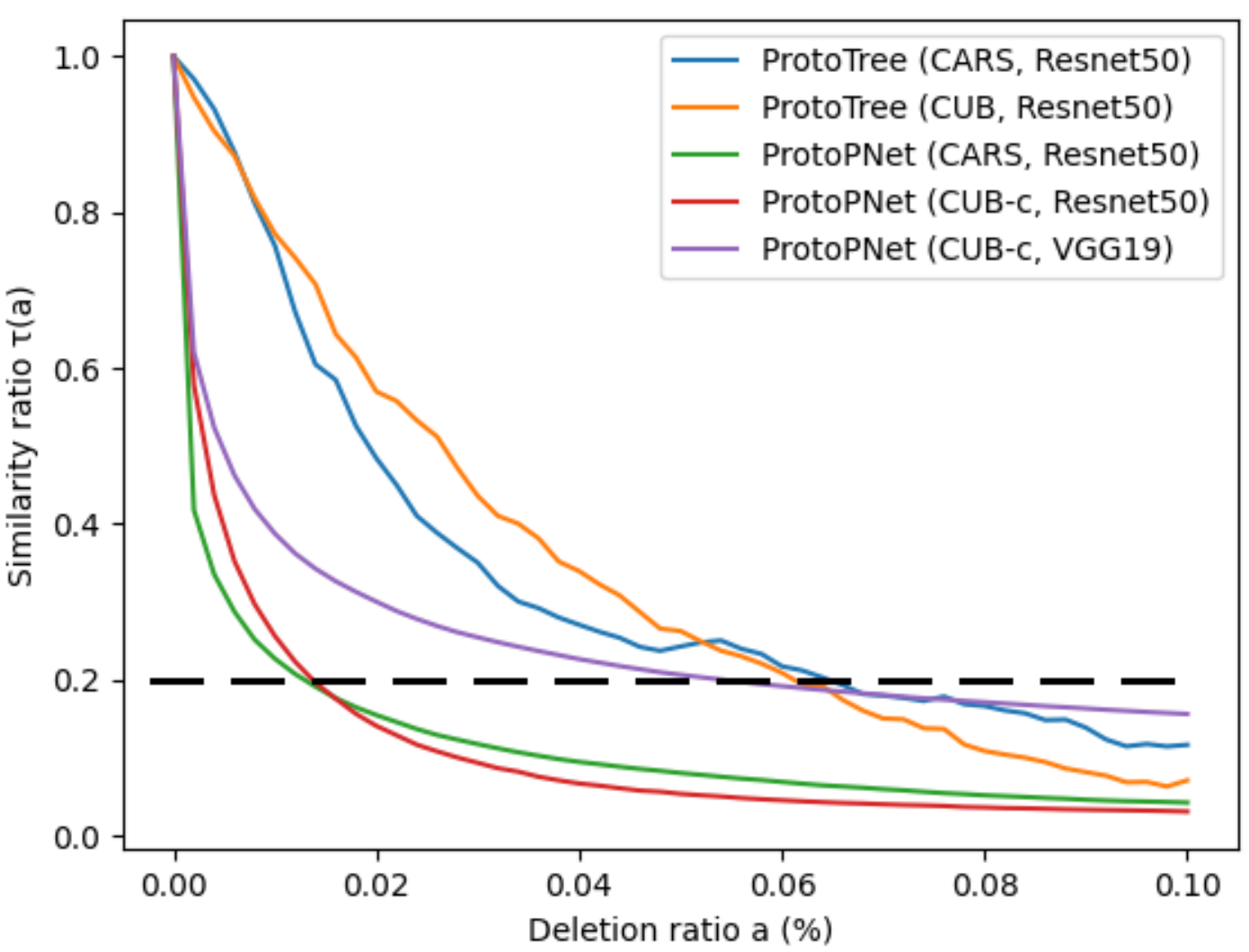}
    \caption{Average similarity ratio v. deletion area when using PRP for visualising prototypes. The average similarity ratio drops "faster" for ProtoPNet prototypes than for ProtoTree, suggesting a greater effective receptive field for ProtoTree prototypes.}
    \label{fig:deletion_curve_prp}
\end{figure}
Interestingly, as shown in Fig.~\ref{fig:deletion_curve_prp}, when extending the deletion area to 10\% of the image, we note that on average, the drop in similarity ratio ($\tau(a) < 0.2$) occurs below 2\% for ProtoPNet prototypes when using ResNet50, and around 7\% for ProtoTree prototypes or ProtoPNet using VGG19. Since the similarity ratio eventually reaches values below $0.2$, this result does not question the faithfulness of PRP explanations. As mentioned in Sec.~\ref{sec:faithfulness}, this effect rather suggests that the size of the effective receptive field of ProtoTree prototypes (or ProtoPNet using VGG19) is larger in general than ProtoPNet with ResNet50, even on the same dataset (CARS). For ProtoPNet with VGG19, this suggests a sensitivity of the model to the underlying backbone, an hypothesis that is reinforced by the next experiment. For ProtoTree, this may be due to the fact that the decision tree shares prototypes among all classes and therefore does not focus on very small details, contrary to ProtoPNet. This effect is also present when using Smoothgrads but seems uncorrelated to the depth of the prototype inside the decision tree (see Supplementary material). Moreover, this clarifies the discrepancy in AUDC scores between ProtoPNet and ProtoTree visualisations. In particular, using the same feature extractor (Resnet50) on the same dataset (CARS), ProtoPNet visualisation with PRP reaches significantly lower AUDC scores (61 for prototypes, 136 for test patches) than ProtoTree (150 for prototypes, 142 for test patches).\\
This first experiment confirms that the upsampling method implemented in ProtoPNet and ProtoTree produces less faithful image patches than PRP of Smoothgrads. Therefore, the explanations provided by default in these models do not necessarily reflect the actual behaviour of the model.

\paragraph{Relevance of patch visualisation}

\begin{table}[]
\renewcommand{\tabcolsep}{0.15cm}
\centering
\scriptsize
	\begin{tabular}{|c|c|ccc|}
		\hline
		\multirow{2}{*}{Model} & \multirow{2}{*}{Backbone} & \multicolumn{3}{c|}{Method} \\
		& & Up. & PRP & $S \odot I$ \\
		\hline
		\multirow{2}{*}{ProtoPNet} & VGG19    & 15.4\% / 23.3\% & 10.\% / 16.6\% & 11.35\% / 19.9\%\\
                                   & ResNet50 & 2.0\% / 8.8\% & 1.3\% / 8.0\% & 1.0\% / 6.1\% \\
        \hline
        ProtoTree & ResNet50 &  35.4\% / 51.9\% & 0.5\% / 0.5\% & 8.7\% / 14.5\% \\
        \hline
	\end{tabular}
	\caption{Percentage of prototypes (left value) and test patches (right value) with less than 5\% of intersection with the object (CUB dataset).}
	\label{tab:relevance}
\end{table}

\begin{figure}
	\centering
    \includegraphics[width=\linewidth]{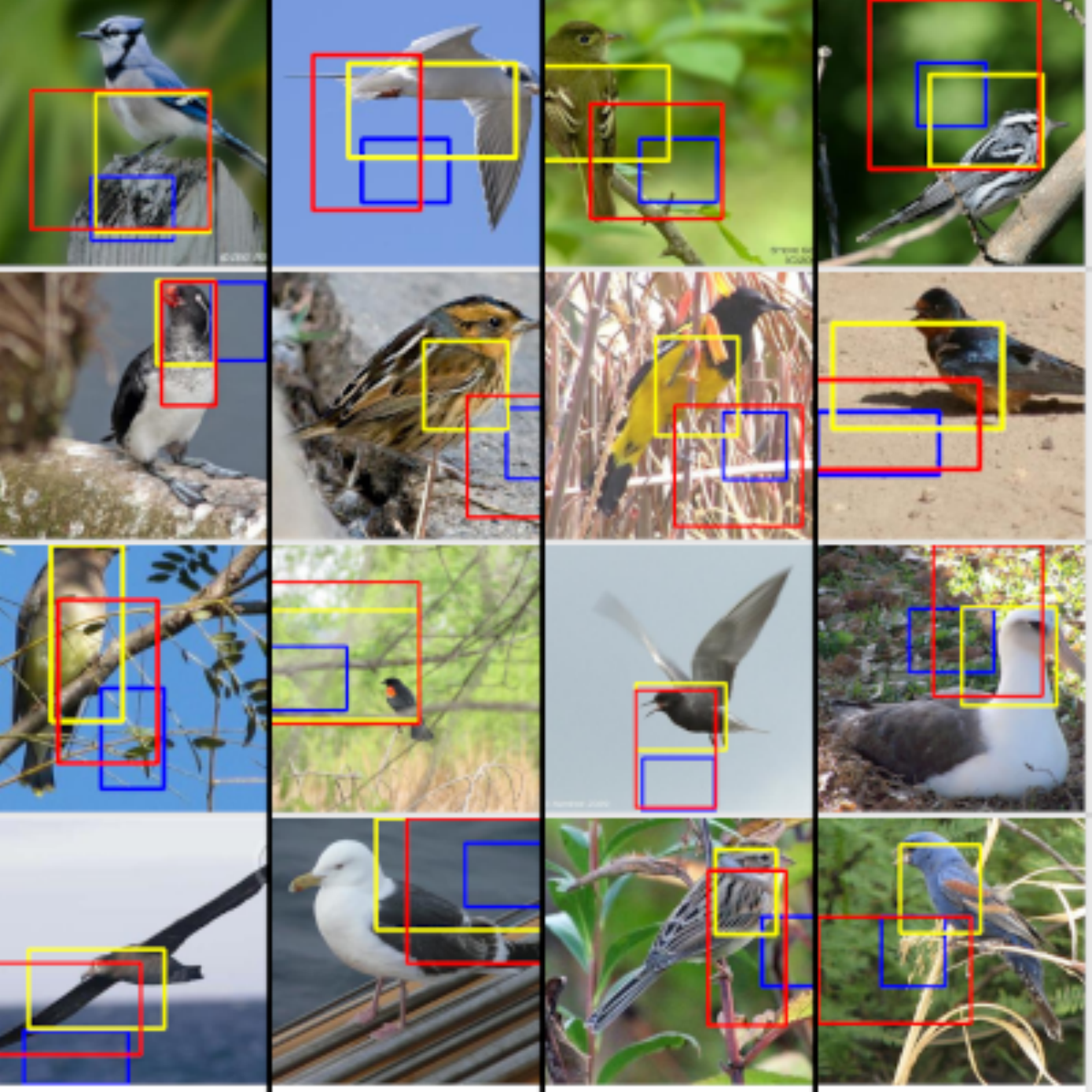}
    \caption{Visualisation of prototypes from a ProtoTree trained on CUB-200-2011 using upsampling with cubic interpolation (blue), Smoothgrads (red) or PRP (yellow). In these examples, the upsampling strategy misses the object, given a false sense of bias in the model. Best viewed in colour.}\label{fig:proto_cmp}
\end{figure}

In this experiment, we use the segmentation masks provided by the CUB-200-2011 dataset (such information is not provided with the Stanford Cars dataset). As indicated in Sec.~\ref{sec:relevance}, we measure the percentage of saliency masks not intersecting the object (intersection of less than 5\%), for both prototypes and test image patches. As shown in Table~\ref{tab:relevance} and illustrated in Fig.~\ref{fig:proto_cmp}, the imprecision of the upsampling visualisation method used in ProtoPNet and ProtoTree leads to a false sense of model bias. In particular, when using the upsampling method, more than a third (35.4\%) of ProtoTree prototypes and half (51.9\%) of the test image patches seem to be focusing on elements of the background rather than the bird. However, when using a more faithful saliency method such as PRP (or even Smoothgrads), we notice that only 0.5\% of the prototypes (\ie a single prototype) and 0.5\% of test image patches are actually irrelevant. Note that this gap between results is again more limited for ProtoPNet where images from the CUB dataset are cropped (in order to achieve a better accuracy) and where the upsampling method is less likely to "miss" the object entirely.
Finally, when applying the PRP method on ProtoPNet, we also notice that the percentage of biased prototypes and test image patches is significantly more important when using VGG19 as a backbone, compared to using Resnet50, which raises the question of the sensitivity of prototype-based architectures to the underlying backbone architecture or to the pre-trained features. \\
In this second experiment, we have shown that the apparent bias of ProtoPNet and ProtoTree suggested by the use of upsampling is largely mitigated when using PRP and Smoothgrads. Far from contradicting the results of~\cite{gautam2022this}, this mainly confirms that the use of an unfaithful saliency method for generating image patches provide unreliable information regarding the actual decision-making process of the model and can be detrimental to the trust in self-explaining models in general.

\section{Discussion and future work}\label{sec:discussion}
Although case-based reasoning architectures for image classification constitute a first stepping stone towards more interpretable computer vision models, such architectures still suffer from several shortcomings that may hinder their widespread usage, especially in critical applications.
In this work, we have shown that even though such models might produce a correct decision for the right reasons (\textit{this indeed looks like that}), they may yet fail to properly explain this decision by incorrectly identifying appropriate parts of the images (prototypes and test image patches). In particular, more faithful saliency methods such as PRP and Smoothgrads can help uncover biases~\cite{gautam2022this} or - in our experiments - disprove \textit{apparent} biases of the model. As indicated in Sec.~\ref{sec:related_work}, this issue is likely not restricted to ProtoPNet or ProtoTree, since ProtoPShare, ProtoPool, Deformable ProtoPNet and TesNet share a common code base inherited from ProtoPNet that includes the upsampling method for patch visualisation.

Additionally, in order to achieve performance (\eg accuracy) on par with non-interpretable methods, case-based reasoning architectures rely on complex feature extractors such as DCNNs, under the assumption that proximity in the latent space entails similarity in the visual space. However, as shown by \cite{hoffmann2021this}, such assumption may not always hold. Moreover, according to our own experiments, the relevance of prototypes and test image patches may vary depending on the choice of backbone architecture (\eg Resnet50 or VGG19), or at least depending on the nature of the pre-trained features used at the beginning of the training process. Finally, proving that the model is indeed focusing on the object to produce its decision does not imply that such decision is based on understandable rather than abstract information. In this regard, although the work of~\cite{nauta2020this} can help shed some light on the visual cues (colour, shape, hue, etc) used to determine similarity, some similarities between image patches remain unintelligible. This suggests that case-based reasoning architectures using DCNNs for feature extraction are not actually explainable-\textit{by-design}, in the sense that a decision-making process based on distance in the latent space is not sufficient to guarantee interpretability. As a consequence, we also argue that raw performance (\eg classification accuracy) should not be used to compare such architectures, as it may drive the research community towards models focusing on more abstract features.

In a future work, we first wish to extend our study to other case-based reasoning models. In particular, ProtoPool~\cite{rymarczyk2021interpretable} uses a focal similarity to ensure the locality of prototypes and to reduce the probability of the model learning elements of the background. Using the metric described in Sec.~\ref{sec:relevance}, this gain in prototype relevance could be quantified. Finally, as stated above, there exists a dire need for metrics for evaluating the understandability of explanations in a systematic manner and, in the case of prototype-based architectures, to properly quantify visual similarity.

\paragraph{Acknowledgments} Experiments presented in this paper were carried out using the Grid'5000 testbed, supported by a scientific interest group hosted by Inria and including CNRS, RENATER and several Universities as well as other organizations (see https://www.grid5000.fr).\\
This work has been partially supported by MIAI@Grenoble Alpes, (ANR-19-P3IA-0003) and TAILOR, a project funded by EU Horizon 2020 research and innovation programme under GA No 952215.

\newpage

{\small

}

\end{document}

\else
\documentclass[runningheads]{llncs}
\usepackage[T1]{fontenc}
\usepackage{graphicx}
\usepackage{amsmath}
\usepackage{amssymb}
\usepackage{subcaption}
\newcommand{\eg}{\emph{e.g.,\xspace}\xspace}
\newcommand{\ie}{\emph{i.e.,\xspace}\xspace}
\newcommand{\etc}{\emph{etc.,\xspace}\xspace}
\newcommand{\wrt}{\emph{w.r.t.\xspace}\xspace}
\usepackage{xspace}
\usepackage{multirow}
\usepackage{booktabs}
\DeclareMathOperator*{\argmax}{arg\,max}
\DeclareMathOperator*{\argmin}{arg\,min}

\newcommand{\etcf}{\emph{etc.\xspace}\xspace}

\newcommand{\rx}[1]{{\color{blue} [RX: #1]}}
\newcommand{\gq}[1]{{\color{green} [GQ: #1]}}
\newcommand{\zc}[1]{{\color[rgb]{1.000000,0.509804,0.278431} [ZC: #1]}}
\newcommand{\mr}[1]{{\color{red} [MR: #1]}}

\newcommand{\feat}[4]{\ensuremath{#1^{(#2,#3,#4)}}}
\newcommand{\featvec}[3]{\ensuremath{#1^{(#2,#3)}}}

\usepackage{hyperref}
\usepackage{color}
\renewcommand\UrlFont{\color{blue}\rmfamily}

\begin{document}
\title{Sanity checks and improvements for patch visualisation in prototype-based image classification}
\titlerunning{Sanity checks and improvements for prototype patch visualisation}

\author{Romain XU-DARME\inst{1,2}\orcidID{0000-0002-8630-5635} \and
Georges QUENOT\inst{2} \and
Zakaria CHIHANI\inst{1} \and Marie-Christine ROUSSET\inst{2}}
\authorrunning{R. Xu-Darme et al.}

\institute{
Université Paris-Saclay, CEA, List, F-91120, Palaiseau, France \\
\email{<first name>.<last name>(at)cea.fr}
\and Univ. Grenoble Alpes, CNRS, Grenoble INP, LIG, F-38000 Grenoble, France \\
\email{<first name>.<last name>(at)imag.fr}
}
\maketitle              
\begin{abstract}
In this work, we perform an in-depth analysis of the visualisation methods implemented in two popular self-explaining models for visual classification based on prototypes - ProtoPNet and ProtoTree. Using two fine-grained datasets (CUB-200-2011 and Stanford Cars), we first show that such methods do not correctly identify the regions of interest inside of the images, and therefore do not reflect the model behaviour. Secondly, using a deletion metric we demonstrate quantitatively that saliency methods such as Smoothgrads or PRP provide more faithful image patches. We also propose a new relevance metric based on the segmentation of the object provided in some datasets (\eg CUB-200-2011) and show that the imprecise patch visualisations generated by ProtoPNet and ProtoTree can create a false sense of bias that can be mitigated by the use of more faithful methods. Finally, we discuss the implications of our findings for other prototype-based models sharing the same visualisation method.

\end{abstract}
\bibliographystyle{splncs04}
\bibliography{../../laiser_pubs/laiser}
\end{document}